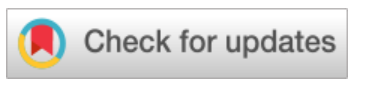

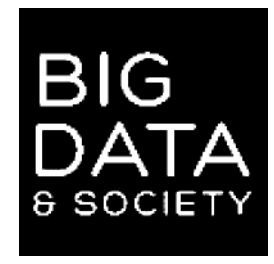

# The pragmatic frames of spurious correlations in machine learning: Interpreting how and why they matter

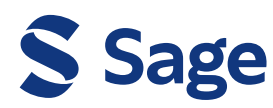

**Samuel J Bell**[1,*] and **Skyler Wang**[2,*]

**Abstract**

Learning correlations from data forms the foundation of today's machine learning (ML) and artificial intelligence research. While contemporary methods enable the automatic discovery of complex patterns, they are prone to failure when unintended correlations are captured. This vulnerability has spurred a growing interest in interrogating spuriousness, which is often seen as a threat to model performance, fairness, and robustness. In this article, we trace departures from the conventional statistical definition of spuriousness—which denotes a non-causal relationship arising from coincidence or confounding—to examine how its meaning is negotiated in ML research. Rather than relying solely on formal definitions, researchers assess spuriousness through what we call *pragmatic frames*: Judgments based on what a correlation does in practice—how it affects model behavior, supports or impedes task performance, or aligns with broader normative goals. Drawing on a broad survey of ML literature, we identify four such frames: Relevance (Models should use correlations that are relevant to the task), generalizability (Models should use correlations that generalize to unseen data), human-likeness (Models should use correlations that a human would use to perform the same task), and harmfulness (Models should use correlations that are not socially or ethically harmful). These representations reveal that correlation desirability is not a fixed statistical property but a situated judgment informed by technical, epistemic, and ethical considerations. By examining how a foundational ML conundrum is problematized in research literature, we contribute to broader conversations on the contingent practices through which technical concepts like spuriousness are defined and operationalized.

## Introduction

Machine learning (ML) forms the backbone of artificial intelligence (AI) research today. Its methods enable researchers to automatically extract patterns from large datasets and make systematic predictions, elevating computationally derived *correlations* to a position of prominence in today's big data era. This new epistemological standard stipulates that computational brute force can stand in for scientific reasoning, thereby diminishing the role of "meaning" while advancing the notion that "numbers speak for themselves" (Calude and Longo, 2017: 595; Kitchin, 2014). Within this framework, causality—the fundamental thread undergirding knowledge production in most scientific fields—becomes secondary as long as one can identify "regularities in very large databases" (Calude and Longo, 2017: 595).

In no subtle terms, journalist Kalev Leetaru (2019) argues that the "entire AI revolution is built on [this] correlation house of cards." Among these correlations, many of them are spurious—they emerge as systems uncover "obscure patterns in vast reams of numbers that may have absolutely nothing to do with the phenomena they are supposed to be measuring" (Leetaru, 2019). An often-cited example of this problem involves training an image recognition model designed to differentiate cows from camels (Beery et al., 2018). Because camels are often

[1]Slingshot AI, London, UK
[2]McGill University, Montreal, Canada

*co-first authorship (equal contribution)

**Corresponding author:**
Skyler Wang, Department of Sociology, McGill University, Room 839, Leacock Building, 855 Sherbrooke Street West, Montreal, QC H3A 2T7.
Email: skyler.wang@mcgill.ca

photographed in the desert and cows in pastures, a classifier trained on conventional photographs of these animals will likely over-index the background color. This leads to models that tend to miscategorize a cow as a camel when the cow is photographed against a yellow background (Beery et al., 2018, via Arjovsky et al., 2019). This issue, appearing under several guises (e.g., "shortcuts, dataset biases, group robustness, simplicity bias"; Ye et al., 2024: 1), demonstrates a typical failure mode of rote pattern-matching systems, whereby spurious correlations are inadvertently leveraged in decision-making.

Spuriousness is a longstanding challenge in ML research. Often depicted as "a major issue" (Volodin et al., 2020), spurious correlations are not only considered "problematic" (Izmailov et al., 2022) but are conventionally regarded as a "threat" to research (Eisenstein, 2022). However, spuriousness is typically investigated only when a model's test set captures failure modes arising from its presence. When this occurs, spuriousness is deemed an impediment to benchmark progress and a problem to be resolved. Yet today's standard ML pipelines remain ill-equipped to distinguish between causation and correlation. This means that when ML researchers deem a set of correlations "spurious," they often bypass the statistical definition of "non-causality" and make sense of the problem through alternative lenses.

How does this sense-making occur? Asking this question compels us to transcend technical considerations of spuriousness and confront the epistemic and normative dimensions of the issue. More specifically, what makes a correlation "spurious," and for what reasons is spuriousness "bad"? Akin to Blodgett et al. (2020), who argue that the way "bias" is conceptualized in natural language processing (NLP) is thin on theoretical engagements, we similarly contend that reducing spuriousness to a one-dimensional (i.e., technical) issue while skirting its underlying epistemic reasonings has troubling implications for ML research.

Take another example—Zech et al.'s (2018) study, which examines the effectiveness of using deep learning models to detect pneumonia from chest X-rays across three American hospitals. Upon examining activation heatmaps of their trained model, the authors found that instead of identifying lung pathologies, the model "learned to detect a metal token that radiology technicians place on the patient in the corner of the image field of view at the time they capture the image" (11). Because technicians from different hospitals place these tokens in slightly different positions, the model uses this "strong feature" to predict disease prevalence. While we recognize the presence of spuriousness (i.e., non-causality) in this case, how should the problem be assessed? Is the model flawed because (1) it picked up a feature that is not *relevant* to the task at hand, (2) it lacks the ability to *generalize* to new contexts (i.e., scans from different patients from different hospitals), (3) it classified in a way that a physician (or *human*) would not, or (4) its potential for misclassification could lead to medical *harm* when deployed in clinical settings?

Through a schematic analysis of ML literature on spuriousness, we find considerable variation in how researchers report when, how, and why such correlations warrant concern. Building on Powell's (2018) work on "operational pragmatics," we argue that rather than relying solely on formal definitions, ML researchers assess spuriousness through what we call *pragmatic frames*—specific ways of seeing that help researchers isolate and interpret aspects of a complex problem. More specifically, pragmatic frames encapsulate the interpretive process by which the validity or undesirability of a correlation is judged not by fixed statistical criteria, but by "the set of justifications made in relation to function" (Powell, 2018: 514)—how it affects model behavior, supports or impedes task performance, or aligns with broader normative goals.

By typologizing four (non-exhaustive) pragmatic frames of spuriousness, *relevance* ("Models should use correlations that are relevant to the task"), *generalizability* ("Models should use correlations that generalize to unseen data"), *human-likeness* ("Models should use correlations that a human would use to perform the same task"), and *harmfulness* ("Models should use correlations that are not socially or ethically harmful"), we demonstrate that the varied paths researchers take to make sense of spuriousness carry distinct technical prescriptions and normative implications. Echoing insights long underscored by Science and Technology Studies (STS) (see Star and Griesemer, 1989; Latour, 1987), we illustrate how a concept like spuriousness becomes meaningful through use, evolving in scope as it becomes entangled in particular research aims and values. As AI systems increasingly permeate social life, we offer this conceptual framework both as a practical tool for building more robust models and a theoretical lens for interrogating the epistemic limits of correlation-based ML.

## Situating the problem: Spuriousness in ML

### Conceptualizing correlation and spuriousness

We begin with some definitional groundwork. Discussing correlation requires disambiguating between two key meanings of the term. In this work, we use "correlation" in its broadest sense to describe any observed relationship between two variables. When we say that variable $X$ correlates with variable $Y$, we mean that a change in $X$ tends to coincide with a change in $Y$. For example, a decrease in daily temperature in Montréal (variable $X$) might correspond to a decrease in the number of people having their morning coffee *en terrasse* (variable $Y$). This relationship could be linear, but we might also expect a plateau (i.e., no matter how warm the weather, at some point every cafe table will be occupied), indicating that it cannot be represented by a straight line. Correlation can refer to any such observed relationship, whether linear or nonlinear, strong or weak, positive or negative. In a second, more narrow sense, correlation may also refer to specific measures of the strength of certain *types* of relationships, such as Pearson's product-moment correlation coefficient for linear relationships.

If correlation describes any relationship observed in the data, then investigating *why* the relationship was observed becomes crucial. The correlation could be causal, meaning that a change in variable $X$ directly causes a change in variable $Y$ (e.g., long periods of drought causing poor agricultural yields). Or it could be non-causal, perhaps arising by chance or due to the presence of a third variable, $Z$, which itself causes changes in both $Y$ and $X$. In statistics and most scientific disciplines, a non-causal correlation of this nature is referred to as a spurious correlation (Pearl, 2009).

In Figure 1, we present several illustrative examples, including causal relationships such as between the moon's phase and the tidal range (Figure 1(b)) or between phase and nighttime luminance (Figure 1(c)). We also highlight spurious correlations: one due to sampling error, between phase and air pollution (Figure 1(e)), and another due to a third variable, between phase and nighttime luminance (Figure 1(f)). This last correlation is spurious because altering the tide would not impact nighttime light directly; instead, the relationship is better explained by a third variable, the lunar phase. Despite being technically spurious, the relationship likely reflects a stable natural phenomenon and, as such, could still prove useful. For instance, if the goal were to predict tide changes, a model using peak nighttime illuminance might fare reasonably well.

If a correlation should be causal in order not to be spurious, then evaluating spuriousness inherently involves the challenge of determining causation. While the nature of causality is a matter of ongoing debate among scientists, a popular view suggests that a causal relationship is one that can be manipulated. That is, a correlation between two variables is causal if intervening upon one variable—changing its value—also results in a change in the other.

Unfortunately, in ML, where models are typically trained on data samples that represent a static snapshot of the world, such interventions are out of reach. Instead, ML researchers must make do with attempts to infer causal structure from observational data. While it should be "in principle possible" (Lopez-Paz et al., 2017: 6980) for algorithms to extract causal structure from natural data such as images, others have argued this is tantamount to a "hopeless task" (Gelman, 2011: 960), incompatible with the fundamental assumptions underpinning standard ML (Schölkopf et al., 2012, Schölkopf et al., 2021), and "only doable in rather limited situations" (Peters et al., 2017: xii). Although causal inference remains a strong and active research area within ML (see, e.g., Peters et al., 2017), today's state-of-the-art ML models rarely benefit from such ideas (Marcus, 2018), likely due to the exceptionally difficult nature of causal inference problems (Peters et al., 2017: xii), a lack of demonstrable advantages (Schölkopf et al., 2021), and the continued success of non-causal alternatives.

## *Machine learning and the "true function"*

When fitting an ML model, developers typically operate with a loose, abstract understanding of a *true function*. That is, given some input data $x$ and a desired outcome $y$, the developer's goal is to find a model that implements some function, $y = f(x)$, that meets their expectations by generally transforming $x$ into $y$. While they may have a clear understanding of *what* the function should do (i.e., transform $x$ into $y$), *how* this should be done is another matter altogether. Precisely defining the intended true function $f$ is sufficiently complex as to be practically impossible (Geirhos et al., 2020), to the extent that researchers may consider the true function to be unknowable without computational support (Goldenfein, 2019). For the ML researcher, this computational support takes the form of optimizing the model to approximate $f$ by identifying correlations within the training data.

More precisely, based on an idea of a function $f$, model developers seek its approximation, $f^*$, typically using an optimization algorithm to minimize error over training data—examples illustrating the behavior of the desired $f$. However, given a fixed training set, there are infinite $f^*$ that will perfectly fit the data, and developers only have limited control over which specific $f^*$ the optimization algorithm will discover. As a result, $f^*$ might be a good approximation of $f$ based on the training data, but it is unlikely to perfectly satisfy the entirety of the developer's expectations—particularly when $f^*$ is tested on rare or unusual samples. Faced with this vast space of learnable functions, developers often introduce informal auxiliary objectives—for example, ensuring that the learned function only relies on a specific subset of available features.

Take, for instance, the aforementioned cow vs. camel recognition case study. For this classifier, the developer defines a function $f$ that, given an image of a cow, outputs the label "cow" and, given an image of a camel, outputs the label "camel." The developer might hope that this function performs consistently across all possible conditions, regardless of how unusual the cow or camel, the weather, the background scene, the image quality, or any other factor may be. To implement such a function, the developer gathers a training set of cow and camel images and uses a learning algorithm to train a model $f^*$ that captures the correlations between features and labels in the training set. However, when tested more broadly, the developer is surprised to find that the cow versus camel classifier fails when presented with images of cows in the desert and camels on grassy pastures. Instead of learning the developer's intended true function, the model has learned to exploit a particular correlation in the training data: cows are often photographed on grassy backgrounds, and camels in deserts. According to the training data, $f^*$ is a good approximation of $f$, but according to the developer's intended function, $f^*$ is no longer satisfactory.

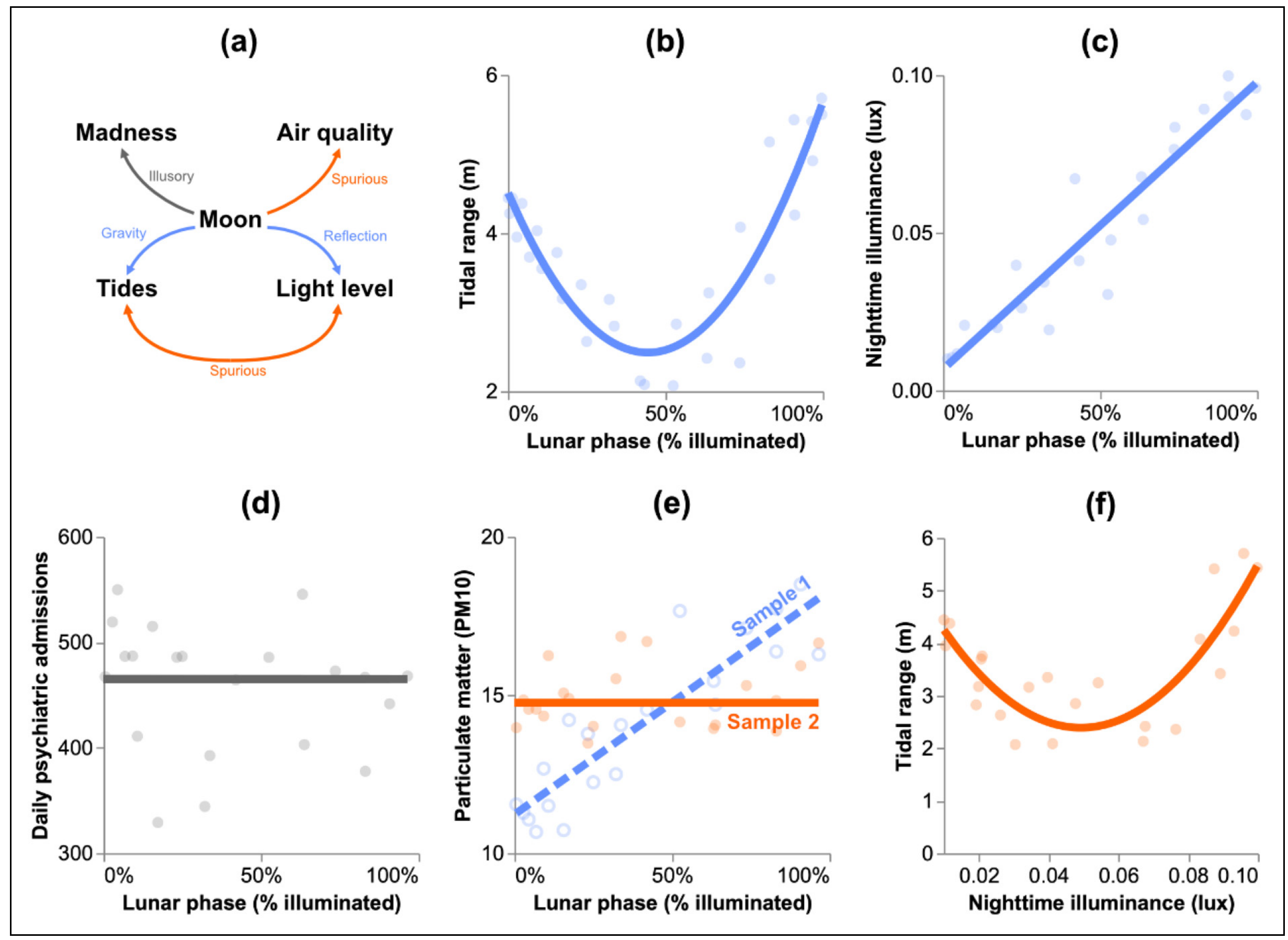

**Figure 1.** Examples of types of correlations. (A) Causal diagram showing different causal, spurious, and illusory correlations. (B) Causal relationship between the moon phase and the tidal range. (C) Causal relationship between moon phase and peak nighttime illuminance. (D) Illusory correlation between moon phase and psychiatric incidence. (E) Spurious correlation due to sampling error between moon phase and air quality, later corrected by a new sample. (F) The predictive, useful correlation between tidal range and nighttime illuminance is spurious because they are both caused by a shared modulating variable, moon phase.

Ultimately, this example perfectly encapsulates the challenge that spurious correlations pose to the modern ML paradigm. While there is a correlation between grass and cows in the training data, and perhaps in the real world, this is not the correlation the developer intended the model to learn. In this sense, the question of which function to learn and which correlations to prioritize is contingent upon the developer's intent and expectations. Given this, what do researchers do when there is a mismatch between intent and reality? In other words, how do ML researchers reconcile the expectation and outcome gap *vis-à-vis* spuriousness?

## *Sense-making and pragmatic frames*

The question of what scientists do when confronted with scientific conundrums is by no means novel. STS scholars have documented a wealth of decision-making, organizational, and cultural tools that researchers deploy to help them overcome uncertainty (Guillaume et al., 2017; Kampourakis and McCain, 2019; Latour and Woolgar, 1979). At their core, these problems create perturbations that compel intellectual reasoning and sense-making, providing us with a window into the social construction of knowledge.

Even though spuriousness has a simple, clear-cut definition in statistics (i.e., non-causal), its negotiation within research is more complex. In our case, when faced with model failures stemming from spuriousness, ML researchers must decide how to represent or frame the problem at hand. As our earlier example illustrates, there are multiple ways one could interpret the "fault" of spuriousness when using ML to detect pneumonia on chest radiograph scans (Zech et al., 2018). Is the model flawed because it did not use relevant features, or because it failed to diagnose the disease as a doctor would? Or is there another underlying issue at play?

To understand how ML researchers navigate ambiguity when defining and problematizing spuriousness, we draw on Powell's (2018) concept of "operational pragmatics"—the practical reasoning and justificatory strategies people deploy when grappling with competing moral, technical, and functional demands, particularly in moments when established meanings are unsettled. Applying this to our case, we analyze how scientific actors construct problem representations in the absence of clear epistemic resolutions. Rather than relying on fixed definitions, researchers mobilize functional reasoning—concerned less with what a concept is in theory than with what it enables them to do in practice. This insight resonates with Knorr's (1979: 352) observation that scientists constantly classify their experience in terms of "what makes sense" and organize their activities accordingly.

In the context of ML, a complex problem like spuriousness can be interpreted differently depending on the task and context. Here, we extend Powell's insight by introducing the concept of *pragmatic frames*: Ways of seeing that help researchers carve complex problems into manageable, actionable parts. These frames guide what kind of problem spuriousness is understood to be, what types of evidence are seen as relevant, and which responses are deemed appropriate. As with "boundary setting" in scientific practice (Hoffmann, 2011; Halffman, 2019; Vazquez et al., 2021), pragmatic framing shapes how failures are interpreted, how solutions are justified, and how resources are allocated. As Entman (2007: 164) puts it, framing involves "culling a few elements of a perceived reality and assembling a narrative that highlights connections among them to promote a particular *interpretation*."

The word "interpretation" is noteworthy in this context. In contrast to the "storybook image of science" (Mitroff, 1974), which stipulates that researchers are detached, objective subjects who deploy standard, positivist instruments to help them map the external world (Mahoney, 1979), STS scholars have long argued that scientific knowledge production is a hermeneutic process where researchers' standpoints, values, and interests fundamentally shape research goals and outcomes. Amidst every effort to find laws in experimental science, there is an interpretative process to find meaning; pragmatic framing is as much about values as it is about scientific accuracy.

Bearing in mind the "social arbitrariness" inherent in research and that, rather than uncovering absolute truths, researchers deploy "tricks" and make a host of decisions to "arrive at" certain truths (Knorr 1979: 347, 352), we assert that how ML researchers pin down a frame—as reflected in how spuriousness is problematized in their publications—may reflect such dynamics. Research priorities are often shaped by personal training, network effects, organizational contexts, and structural incentives (Wang et al., 2024). A lab with social science expertise may be more inclined to investigate how "human" something is, whereas a lab lacking such expertise may prioritize other ways of situated seeing. In short, framing spuriousness pragmatically transforms an abstract conundrum into a solvable task—one that inadvertently reflects the contingencies in a researcher's goals, constraints, and values.

## The pragmatic frames of spuriousness

We explore the pragmatic frames of spuriousness in ML research through a two-stage, critical literature analysis. In the first stage, we used the Semantic Scholar API to retrieve research publications matching the search terms "spurious correlation," "spurious feature," and "shortcut," with at least one citation and an open-access PDF available. Our search included preprints, journal articles, and conference papers in ML, NLP, and computer vision venues, yielding 713 papers. From these, we selected a subset of 200 papers via uniform random sampling for manual screening. Irrelevant papers (e.g., those referring to architectural "shortcut connections") and low-quality preprints were excluded, resulting in 49 relevant papers. By reviewing bibliographies, we further identified 16 frequently cited and substantively relevant works, bringing the first-round analytical sample to 65 papers. These papers span work that defines correlations, biases, and shortcuts, as well as studies focused on mitigation and robustness.

This first round was used to surface preliminary analytic codes and identify recurring ways in which spuriousness is framed and justified in the literature. Both authors independently reviewed and coded these 65 papers, extracted supporting quotations, and then met to compare interpretations and iteratively refine a shared codebook. Notably, we documented very few papers that relied on causal theory or formal causal definitions to characterize spuriousness. In the absence of such definitions, spuriousness was instead articulated through four dominant pragmatic frames: Relevance, generalizability, human-likeness, and harmfulness. Each paper in the sample was assigned one or more frames or marked as missing a definition during a round of interactive discussion.

In the second stage, we returned to the remaining 513 papers that were not included in the first-round close reading. With the assistance of a trained research assistant, we first screened out irrelevant papers and low-quality preprints. We then applied the established codebook to the remaining papers, while being attentive to potentially novel rationales or interpretations of spuriousness that did not fit the existing categories. As we examined the broader corpus, we found that spuriousness could be consistently interpreted using the four pragmatic frames developed in the first round, suggesting that these frames capture the dominant interpretive logics in existing ML literature. This second round yielded 222 additional articles, bringing the total number of papers in the final sample to 287 (see Supplemental Materials, Table S1, for the complete list). Table 1 summarizes the typology, along with typical rationales and illustrative examples for each frame. Across the full sample, the most prevalent frame was generalizability, followed by relevance, harmfulness, and human-likeness (see Table 2).

**Table 1.** The pragmatic frames of spuriousness as distilled from Machine Learning (ML) publications.

| Frame | Typical Rationale | Illustrative Example |
|---|---|---|
| Relevance | Models should only use correlations that are relevant to the task. | "Deep classifiers are known to rely on spurious features—patterns which are correlated with the target on the training data but **not inherently relevant to the learning problem.**" (Izmailov et al., 2022: 1) |
| Generalizability | Models should only use correlations that generalize to unseen data. | "A key challenge in building robust image classification models is the existence of spurious correlations: misleading heuristics imbibed within the training dataset that are correlated with majority examples **but do not hold in general**." (Ghosal et al., 2022: 1) |
| Human-likeness | Models should only use correlations that a human would use to perform the same task. | "A shape-agnostic decision rule that merely relies on texture properties clearly **fails to capture the task of object recognition as it is understood for human vision.**" (Geirhos et al., 2020: 668) |
| Harmfulness | Models should only use correlations that are not harmful. | "Empirical work has shown that deep networks find superficial ways to predict the label … Such behavior is of practical concern because **accuracy can deteriorate under shifts in those features** … It can also lead to **unfair biases and poor performance on minority groups.**" (Nagarajan et al., 2020: 3) |

**Table 2.** Counts indicating how many papers invoke each pragmatic frame.

| Classification | Number of Papers* | Percentage (out of the Total 287 Papers) |
|---|---|---|
| Relevance | 104 | 36% |
| Generalizability | 126 | 44% |
| Human-likeness | 53 | 18% |
| Harmfulness | 80 | 28% |
| Casual | 12 | 4% |

Because frames are not mutually exclusive, a single paper may be coded under multiple categories, and totals therefore exceed the number of unique papers ($N = 287$).

We note that while researchers occasionally explicitly define spuriousness in the exact terms we landed on, they more often leave it loosely stated and use these pragmatic frames as motivations for developing solutions to address it. When doing so, many appear to overemphasize certain frames while omitting other candidates. Although "spurious" literally means incorrect or fallacious, the frames primarily reflect developers' context-dependent and often conflicting demands regarding their systems' behavior.

While our analysis is based on published research papers rather than ethnographic observation of researchers or labs, we argue that examining how researchers frame problems and justify choices in written outputs—the primary artifacts through which knowledge circulates—offers a valuable lens into the epistemic cultures of a field. For most readers, including other scientists, policymakers, and the public, these publications are one of the few points of access to the inner workings of ML research. Attending to what gets foregrounded or routinized in these texts reveals not only what researchers think but also what they deem worthy of communication and legitimation.

Below, in addition to analyzing each of the four frames, we also offer critiques of these frames within the context of ML publications. Our critique is meant to be generative: By acknowledging the gap between statistical definitions and how spuriousness is actually interpreted, we point to the need for methodological tools that attenuate the effects of spuriousness by working with, rather than ignoring, these pragmatic frames. Ultimately, we see these frames not only as diagnostic tools but also as resources for improving future ML methods and honing new research agendas. As we will demonstrate, what makes a correlation spurious is rarely a property of the data itself; it is more often a reflection of the developer's or user's intent or expectations. Spuriousness, it seems, is in the eye of the beholder.

## *Relevance*

Modern ML methods excel at automatically extracting relationships from raw, unprocessed data such as images, text, or audio. Often regarded as a key strength of deep learning,

"automatic feature extraction" refers to how models can learn to hierarchically transform low-level inputs, such as pixels, into increasingly complex, higher-level, and "task-relevant" representations (Goodfellow et al., 2016). This automatic approach has fully supplanted the time-consuming and error-prone practice of "feature engineering," in which developers would write code to transform raw input data into something appropriate for the model. In this manual process, developers would explicitly state their assumptions—and indeed their expertise—about what aspects of the data were relevant to the task and should be used by the model. In contrast, with automatic feature extraction, models are trained to infer which features are relevant to solving the given learning objective based on the samples observed during training.

This shift, however, comes with a trade-off: the correlations the model considers "relevant" (i.e., the relationships it learns to leverage) do not always align with those a developer might deem germane or a feature engineer would have chosen to extract. For example, Reimers et al. (2021: 2) consider "spurious dependencies" to be those relying on "meaningless features." Indeed, contemporary models may have an inductive bias towards "simpler" decision rules (Arpit et al., 2017; Kalimeris et al., 2019), which can lead models to interpret trivial artifacts as relevant features (Shah et al., 2020). When the model-relevant features diverge from developer-relevant features, researchers often label the correlation as spurious. In other words, researchers expect their models to use only the correlations they consider relevant.

Izmailov et al. (2022: 1) make this view apparent, stating that spurious correlations are "not inherently relevant to the learning problem," while Ghosal et al. (2022: 2) claim that spurious features are "statistically informative … but do not capture essential cues related to the labels." Joshi et al. (2022: 9804) describe a correlation as "undesirable" if "the feature is irrelevant to the label" and "neither necessary nor sufficient for prediction." Similarly, Kirichenko et al. (2023: 1) describe spurious features as "predictive of the target in the train data, but that are irrelevant to the true labeling function," highlighting the clear role of researcher expectations: Spuriousness via relevance is defined in relation to the idealized "true labeling function." A common thread in these examples is the perspective that a learned but spurious correlation may be relevant within the training data but becomes less relevant in a more general context or under a different data distribution. Scimeca et al. (2022: 1) note that deep neural networks "often pick up simple, non-essential cues, which are nonetheless effective *within a particular dataset* [emphasis added]." In this way, the relevance frame establishes a multipartite relationship among the researcher's expectations for the intended task, the training data, and possible future data distributions (a subject we will revisit in "Generalizability").

Perhaps the most straightforward example of relevance assumptions occurs in computer vision, where it is commonly assumed that image foregrounds are always pertinent to the task, while background features are deemed irrelevant and, therefore, spurious. For instance, Ye et al. (2024: 1) consider "background, texture, and secondary objects" to be "non-essential features" and any resulting correlations to be spurious. Also adopting an object-centric definition of relevance, Singla and Feizi (2022) find that flowers commonly co-occur with images of insects in the popular image classification dataset ImageNet (Deng et al., 2009) and suggest that flowers are, therefore, spurious because they are not a "part of the relevant object definition [of insects]." However, "relevant" unfortunately remains undefined. In constructing a dataset to test for susceptibility to spurious correlations, Sagawa et al.'s (2019) Waterbirds dataset assumes that background is irrelevant to distinguishing water-dwelling from land-dwelling birds. The same judgment is at play in the classification of cows and camels as mentioned above, or the well-known case of computationally distinguishing huskies from wolves (against snowy and non-snowy backgrounds, respectively; Ribeiro et al., 2016).

These assumptions might make sense in the abstract, formalized setting of image classification commonly practiced in ML, where developers compete to create models that assign labels to images of objects. In its most common form, models are trained to assign just a single label to each image, drawn from a predefined set of available classes. However, many, if not most, images are not of single objects but complex scenes comprising multiple objects in context. For example, the image in Figure 2 is taken from the "tench" class of ImageNet, one of 1000 possible classes including "scuba diver" and "sturgeon," but not "angler" or "person." In this single-label setting, "tench" is the only label to be applied, meaning the only relevant, non-spurious correlations should be those tied to the fish pixels and label. Yet, to fully understand the image, one must recognize that it clearly contains more than just a tench: It shows a person who has caught a fish, is proud of their catch, and is posing for a photo to commemorate the moment. Thus, if "tench" features are considered the only relevant ones, they reflect the constraints of the available labels and the narrow horizons of image classification rather than the intrinsic qualities of the image itself. Relevance judgments, therefore, arise from the researchers' interpretations of the intended task and application rather than from the inherent properties of the data.

## Generalizability

The canonical method for evaluating an ML model is to test its predictive performance on unseen (or "held-out") data. This process is designed to assess whether the model can generalize what it has learned from the training samples to a "test set" of samples assumed to be drawn from the same distribution. During training, models are optimized to extract and leverage correlations that explain the training data, but only a subset of these correlations will be useful for explaining the test data. Concerning generalizability, correlations that do not hold at test time are typically described as spurious, and any model that relies on them is said to have overfit the training data. Thus, we define the second pragmatic frame of spuriousness as generalizability, as researchers expect models to learn correlations that generalize beyond the training data.

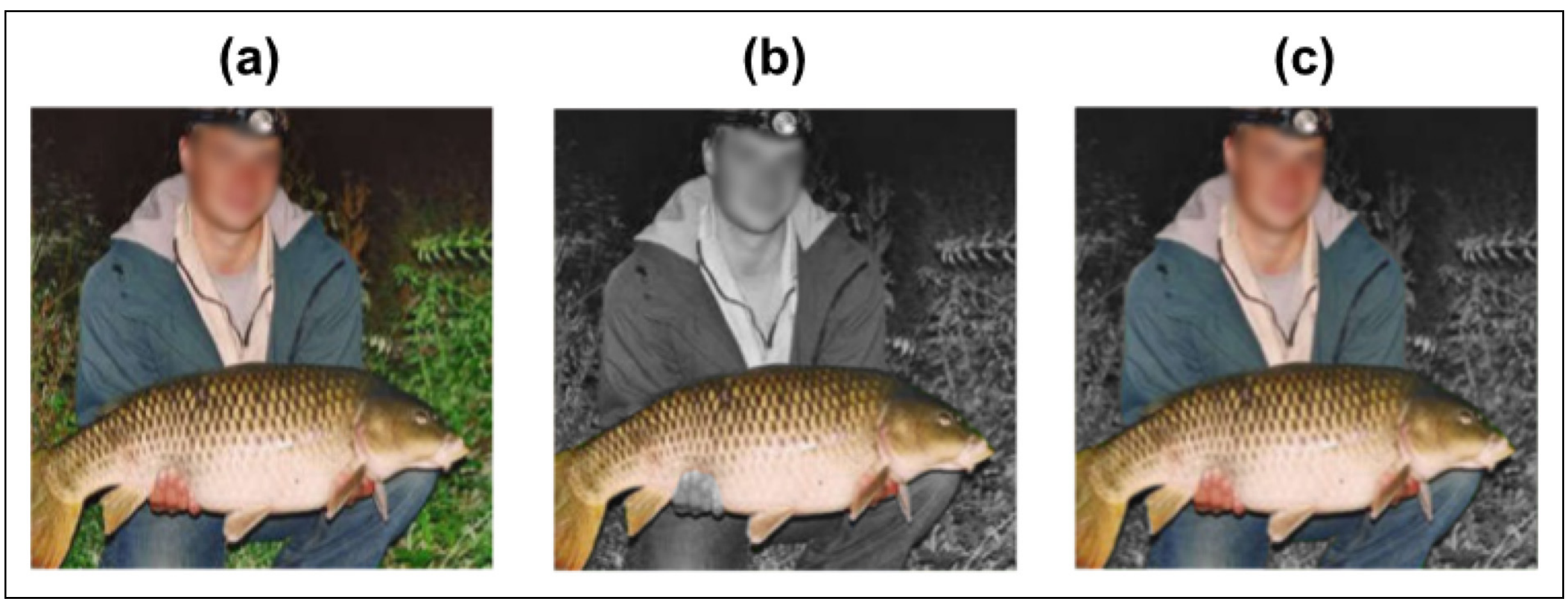

**Figure 2.** (A) Example image of class n01440764 "tench" from ImageNet. (B) Only the pixels of the tench itself pertain to relevant, non-spurious correlations in 1000-class image classification. (C) An alternative interpretation of relevance, perhaps as understood by the photographer or subject. Faces blurred for privacy preservation.

Turning to the literature, Nagarajan et al. (2020: 1) describe how models often leverage features that are "spuriously correlated with the label only during training time, resulting in poor accuracy during test-time." Similarly, Yang et al. (2023: 3) define spurious correlations as those that hold in "training but not in test data," while for Lu et al. (2022: 1), they "fail to generalize to environments whose distributions differ from the ones used at training time." Joshi et al. (2023: 1) describe spurious features as those which are "correlated with a given class, but are not predictive of class membership." Also concerned with prediction, Benoit et al. (2023: 1) see spurious correlations as those which "lose their predictive power under a distribution shift and consequently fail to generalize to out-of-distribution." These examples clearly illustrate the salience of generalizability when assessing spuriousness in ML publications.

Increasingly, ML researchers apply a more stringent test of generalization, constructing out-of-distribution test sets in which samples are drawn from meaningfully different distributions. These "stress tests" (Lopez-Paz et al., 2022) assess not only how the model performs on unseen test data but also how it handles important conditions that may be completely absent from both the training data and standard test sets. For example, researchers may test whether an image classifier trained on photographs of objects can generalize to (i.e., perform equally well on) hand-drawn sketches of the same set of objects (Wang et al., 2019a). Unfortunately, it is rare for models to achieve such generalization; instead, they tend to learn correlations that hold in the training set—and perhaps in the test set—but not in out-of-distribution settings. Framed through concerns around generalizability, these correlations are typically considered spurious.

Researchers often explicitly state that correlations must be general to avoid being spurious. For example, Ghosal et al. (2022: 1) describe spurious correlations as "misleading heuristics within the training dataset … [that] do not hold in general." Similarly, Arjovsky et al. (2019: 1) argue that correlations must be "stable," and a correlation is spurious if "we do not expect it to hold in the future in the same manner as it held in the past." Pezeshki et al. (2024: 1) refer to spurious correlations as "environment-specific," meaning they only hold under a subset of contexts likely to be encountered, while Sreekumar and Boddeti (2023: 1) argue that for a correlation to be spurious, it must "not hold under *natural* distribution shifts [emphasis added]." Geirhos et al.'s (2020: 665) investigation of how models learn shortcuts, decision rules that rely on spurious correlations, suggests that the challenge is to "transfer to more challenging testing conditions, such as real-world scenarios." Here, we see that generalizing to the test set is no longer sufficient; instead, correlations must generalize much further to other data distributions and unseen contexts.

It is here, in justifying correlations through their generalizability to other data distributions, that we observe a critical omission: researchers rarely specify to *which* distributions and contexts the function should generalize. As we observed with the relevance frame, this gap again highlights the contingent ways in which researchers reason about correlations. At a fundamental level, a function cannot generalize to *all* possible data distributions. By analogy with Wolpert and Macready's (1997) *no free lunch theorem,* this requirement would render all correlations spurious, as all possible distributions must include those where the correlation no longer holds. Instead, while appealing to generalizability, researchers seek correlations that generalize in specific ways. For example, taking a causal approach, Arjovsky et al. (2019: 10) expect correlations to generalize to distributions that can be constructed via "valid interventions" (i.e., those that modify the data's causal structure without changing the outcome variable). Given that the underlying causal graph is almost always unknown, the practical design and implementation of such interventions remains a pivotal challenge. Alternatively, researchers may specify properties of appropriate test distributions, such as Sreekumar and Boddeti's (2023: 1) "*natural* distribution shifts" and Geirhos et al.'s (2020: 665) "*real world* scenarios," yet such broad descriptions belie the fact that researchers must still decide on concrete instantiation. Taking an image classification example, whether a natural shift is interpreted as images from a different city or a different continent will modulate whether the correlation generalizes.

In defining stress tests, researchers must first reasonably characterize an unseen data distribution and then translate it into a set of samples for testing. This challenging process is naturally contingent on both understanding future data distributions and the ability to practically implement test sets that are sufficiently representative of those distributions. Recognizing the need for diverse viewpoints, Lopez-Paz et al. (2022) suggest that stress tests be defined by parties with differing interests, including model developers and intended users. By specifying expectations about how models should generalize in their publications, researchers implicitly encode which aspects of the data matter and, by extension, determine which correlations are valid and which are spurious.

## Human-likeness

Since its inception, AI research has sought to emulate certain aspects of human intelligence, including both human-like behavior and the mechanisms underlying it. Therefore, it's perhaps unsurprising that when ML models fail to meet certain predefined expectations, a common response is to suggest they are not sufficiently "human." Under the human-likeness frame, relationships between features a human would not use are readily described as spurious, and models that leverage such spurious correlations may be seen as failures.

For example, in their highly cited work exposing the propensity of neural networks to learn "shortcuts" (a term often considered synonymous with spurious correlations), Geirhos et al. (2019) point to how contemporary models overly rely on texture-based cues (such as an animal's fur or skin) instead of those based on shape. Appealing directly to human-likeness, Geirhos et al. (2020) cite evidence about the well-known human bias toward object shape to argue that a "shape-agnostic decision rule that merely relies on texture properties clearly fails to capture the task of object recognition as it is understood for *human* vision" (668; emphasis ours). A human-centric understanding of the task is also highlighted by Yue et al. (2023: 1), for whom spurious correlations "would not be identified as rationales to the prediction task by human." Similarly, Scimeca et al. (2022: 1) critique how "shortcut biases often result in a striking qualitative difference between human and machine recognition systems." McCoy et al. (2019: 1) highlight a gap between human and machine generalization, arguing that "shallow heuristics" provide "little incentive for the model to learn to generalize to more challenging cases as a human performing the task would." Explicitly turning to human judgments, Srivastava et al. (2020: 1) propose "making models robust to spurious correlations by leveraging humans' common sense knowledge of causality." In each of these examples, learning to use human-like correlations—correlations that a human uses for inference—remains the underlying goal.

Take another example: In contemporary ML discourse, complex human faculties such as visual perception are often reduced to computational problems to be solved (Denton et al., 2021). Efforts to "solve vision" entail developing ML systems that respond in an adequately human-like fashion across a range of tasks. Such an ambition, however, presupposes that we are in possession of a complete, accurate, and universal account of human visual perception and cognition—a presumption that is far from reality. For instance, over the last fifteen years, the psychological sciences have reckoned with growing evidence that many behaviors previously thought to be universal do not, in fact, generalize. The vast majority of human psychological research is conducted within and draws participants from Western, Educated, Industrialized, Rich, and Democratic (WEIRD) societies (Henrich et al., 2010). Even within the WEIRD sample, participant pools tend to be dominated by college-aged women (Gosling et al., 2004). Henrich et al. (2010: 4) report that key findings in the domain of visual perception show "substantial differences" when retested on samples drawn from other societies. As these authors note, cross-cultural differences do not negate the existence of universal human behaviors but pose a significant challenge to identifying which behaviors are shared and which are culturally specific. If a computer model must learn a function corresponding to human vision, the variation in human vision across cultures makes this a highly subjective and contingent demand.

This quest for human-likeness extends beyond the behavioral to the mechanistic, such that it is often considered insufficient for a model to merely learn a function whose behavior matches that of a human. Instead, for researchers who care about this pragmatic frame, the model must also exhibit its behavior in a human-like manner and employ human-like *internal mechanisms*. In other words, not only should the learned function produce identical outputs to a human's, but it should also, in a deeper sense, be equivalent to the human function. However, claims of implementational equivalence between human-learned and machine-learned functions are necessarily scoped by analysis resolution. At base, the functions cannot be identical: The former relies on biological neurons, while the latter is implemented on a digital machine. A more charitable interpretation of mechanistic similarity might focus on the cognitive level rather than the neurobiological one. Yet, claims that artificial functions are composed of the same cognitive building blocks as human functions would require a stable, generalizable, and robust understanding of which building blocks humans use. Absent such an understanding, almost any function could be arbitrarily described as either human-like or spurious.

Whether or not we understand or can emulate a human-learned function sidesteps a crucial question: Is it truly desirable to do so? While researchers might judge the acceptability of a correlation through the human-likeness frame, this may conflict with society's—or even the same researchers'—broader aims for AI. In many domains, the explicit goal is to develop systems that surpass human performance on critical tasks, as evidenced by the slew of papers claiming superhuman performance in medical diagnostics (e.g., McKinney et al., 2020; Yim et al., 2020) or by the stated ambitions of the numerous AI companies promising to achieve superhuman intelligence (Edwards, 2024). Additionally, we may often desire systems that operate in fundamentally different ways to circumvent the fallible, biased decision-making exhibited by humans. Unlike the relatively straightforward interpretation of spurious as non-causal, an interpretation of spurious using the human-likeness frame is once again task-dependent and context-specific.

### Harmfulness

ML models have repeatedly been shown to learn, replicate, and amplify potentially harmful biases in their training data (Bolukbasi et al., 2016; Hendricks et al., 2018; Stock and Cissé, 2018; Zhao et al., 2017). This occurs when models learn correlations involving socially sensitive or protected attributes, such as race or gender (even when they are deliberately instructed not to), leading to biased decision-making. A well-known example is Amazon's automated candidate screening tool, which downranked the résumés of women applicants because the résumés it was trained on were predominantly from men (Dastin, 2018). Consequently, models may learn to leverage correlations that hold only for a subset of the data, resulting in performance disparities when applied to other groups or under different conditions. Many common failures can be ascribed to this issue, such as voice assistants that fail to respond to the accents of minoritized speakers (Harwell, 2018) or autonomous vehicles that struggle to recognize darker-skinned pedestrians (Li et al., 2024). From these examples and the extensive body of research documenting bias in data and in models, we can infer that researchers seek models that avoid causing harm. In other words, when framed in terms of harmfulness, they expect their models to learn functions that steer clear of harmful correlations.

In their publications, ML researchers frequently describe correlations and the models that use them in relation to harm. For example, Wang et al. (2019b: 5310) argue that models that are "sensitive to spurious correlations … risk amplifying societal stereotypes," while Limisiewicz et al. (2023: 1) consider spurious correlations themselves as "stemming from biases and stereotypes present in the training data." Scimeca et al. (2022: 9) claim that "relying on simple cues is sometimes unethical" and describe it as an "alarming phenomenon" (where "simple" in this context implies spurious). Several authors discuss the resulting failures, such as Nagarajan et al.'s (2020: 3) suggestion that relying on spurious correlations can "lead to unfair biases and poor performance on minority groups," or Hartvigsen et al.'s (2022: 1) assertion that an online toxicity detection system's "overreliance on spurious correlations" leads to a disproportionate number of false positives for minoritized groups.

As seen above, researchers use various criteria to determine whether a correlation is harmful. In some contexts, only correlations that cause representational harm (sometimes referred to as "harmful associations"; Goyal et al., 2022) are undesirable, whereas, in others, model developers may need to avoid any correlation involving a sensitive or protected attribute, regardless of whether it leads to representational or allocative harm. For example, CelebA (Liu et al., 2015), a popular dataset of celebrity faces commonly used in spurious correlations research, exhibits a correlation between gender and hair. In CelebA, 95% of images of people with blond hair are women, whereas only 5% are men. As a result, models trained to classify hair color tend to also infer and use gender. This particular correlation is deemed spurious because it relies on gender, a sensitive or protected attribute, yet it is unlikely to cause significant representational harm.

Alternatively, a distinct form of harm arises from correlations that are not inherently harmful but lead to model failures when the correlation no longer holds, or cause performance disparities when it holds only for certain groups. In toxicity detection, it is a true, stable, yet unfortunate fact that many minoritized groups are overwhelmingly the subject of online hate and abuse—that is, there is a non-spurious correlation between mentions of minoritized groups and toxicity. However, if the learned function leverages this correlation, the model will penalize innocent mentions of those groups, resulting in harm in the form of performance disparities (Dixon et al., 2018) or through the disproportionate impacts of content moderation. For example, Dias Oliva et al. (2021) find that online toxicity detection systems often classify neutral identity markers such as "gay" or "lesbian" as highly toxic, potentially leading to the over-censorship of LGBTQ + digital content.

In each of these examples, avoiding harm depends not only on understanding which correlations the data capture but also on deciding which relationships to prioritize. This is a far more complex question. A naive query for a "nurse" or a "CEO" with a contemporary generative text-to-image model might return mostly images of uniformed women in hospital environments for the former and besuited men in boardrooms for the latter, reflecting a training dataset where most nurses are women and most CEOs are men. Indeed, globally, most nurses are women, and most CEOs of S&P 500 companies are men; therefore, these correlations in the training data correspond, in some sense, to stable facts about the world. Yet, a model returning such results could rightly be criticized for exhibiting gender bias and reinforcing stereotypes. If researchers use harm as their pragmatic justification of spuriousness, then what matters is how they choose to represent the world: as it exists today, where occupations remain gendered, or as it could be in a future where these roles are perhaps less constrained.

These judgments—about how to represent the world and which relationships models should learn—extend beyond what is in the data and how effectively a model learns to represent that data. Harmfulness, like the other pragmatic frames of spuriousness, requires a clear definition of the function the developer intends the model to learn and which correlations are deemed permissible in the process.

## Discussion and conclusion

The success of modern ML systems hinges on their ability to automatically extract and use correlations found in large datasets. This strength, however, comes with two fundamental challenges. First, as datasets grow ever larger, it becomes increasingly difficult to understand which correlations are encoded within them. Second, given the infinite number of possible correlations one could extract from any dataset, the learning algorithms used to train models must naturally prioritize certain correlations over others. Together, these challenges lead to models that can automatically learn from data structure, but, in reality, often learn functions over which researchers have limited control, resulting in models that leverage the "wrong" correlations. While these bad correlations may be readily described as spurious, modern ML lacks effective tools to distinguish causal from non-causal relationships (Schölkopf et al., 2021). Sidestepping a definition rooted in causality, we argue that researchers instead tend to assess the acceptability of correlations through pragmatic frames, or a set of justifications used to problematize spuriousness. In this article, we have identified four such frames through a scoping review of ML literature: Relevance, generalizability, human-likeness, and harmfulness.

While each frame may support varied reasoning about spuriousness vis-à-vis correlation desirability, each also presents its own challenges. To determine whether a correlation is relevant, one needs a clear specification of the task the model intends to solve. In contrast, assessing generalizability requires reasoning about what data distributions are likely to be encountered—or, perhaps more importantly, which ones matter. Similarly, the human-likeness frame relies on a stable understanding of human functioning, whereas the harmfulness frame hinges on a specific definition of downstream harms to be avoided. By skirting causality and instead leveraging the four pragmatic frames explored in this article, researchers implicitly transform the problem of spurious correlations from a static property of the data into a dynamic relationship between the data and the researcher's expectations.

While researchers might reason about the acceptability of correlations through various pragmatic frames, they often describe correlations in terms of spuriousness, albeit without explicitly considering causality. This rhetorical trick serves to obfuscate the fact that correlation acceptability is highly contextual: In certain situations, it may be desirable to have a human-like function, while in others, relevance might be a more appropriate frame. Research on spurious correlations often fails to account for this contingency, resulting in scientific progress that may not always translate into practical downstream applications (Bell et al., 2024). While the ML community continues to produce swathes of research on theorizing, evaluating, and mitigating the effects of spurious correlations (narrowly understood), reasoning about which functions should be learned remains a challenging and underexplored area. Fully integrating a more explicit understanding of why some correlations are deemed "good" while others are "bad" might enable the development of new approaches for both benchmarking and instilling desirable model behavior. This could, for example, result in new benchmarks for human–machine comparison or methods for explicitly reducing harm rather than reducing over-reliance on spurious features.

Ultimately, the lack of consistent language around spuriousness may directly impact how the ML communities organize themselves in response to this type of model failure. Recognizing that concepts may "contain multitudes" (Whitman, 2005: 43) and face tensions between "coherence and polysemy" (Navon, 2024: 21) in interpretation, we offer these pragmatic frames of spuriousness to theorize varied research priorities under a unifying lens. However, we want to stress that the frames delineated in this article are not exhaustive. As extant STS literature illustrates, scientific concepts are rarely static. Concepts can change how they get "represented, used, or acted upon" (Navon, 2024: 21) as they interface with shifting epistemic environments and contexts. As ML research evolves to cover increasingly multimodal and agent-based areas of interest, and as advances in training paradigms and model architectures reshape how systems learn and generalize, we may encounter new framings of spuriousness that are not explored in this current exposition.

In their book *The Ordinal Society*, Fourcade and Healy (2024: 2) cogently note that even when we know that the "data is bad" and the "results are spurious," ML-driven systems are often cloaked in a veneer of rationalism and neutrality. By shining a spotlight on the sense-making logics and interpretative nature of ML problem-solving via its publication outputs, we contribute another key piece of evidence uncovering the "contingent epistemic assumptions, choices, and decisions" of a field long associated with dispassionate objectivity (Denton et al., 2021: 3). Our analysis further underscores the interconnectedness of epistemic and normative choices. When an ML researcher prioritizes one pragmatic frame over another, they make a value-laden decision that comes with trade-offs. For instance, over-indexing on relevance rather than human-likeness could pose greater risks if a system is designed to operate in settings where human decision-making is the gold standard. However, if researchers and developers choose to forego the human-likeness frame because its mechanisms require more time and resources to ascertain, that decision should be viewed as both epistemic and normative, implicating scientific and moral accountability.

Finally, by reconsidering how ML researchers make sense of a common research problem, we hope to encourage future research into how other, equally multifaceted concepts are interpreted in ML contexts. Interrogating the practical norms governing how proverbial research knots get untangled can foster more reflexive and robust epistemic environments, potentially leading to greater investments in fairness and interpretability research. Through systematized efforts, we may find ourselves just fortunate enough to inhabit a world where stronger, more social-centered models (Wang et al., 2024) accessible to people across cultural and institutional contexts become the norm rather than the exception.

## Acknowledgments

We are indebted to Levent Sagun for their contributions to the project's conceptualization and early development. This article benefited greatly from numerous insightful conversations with Diane Bouchacourt, David Lopez-Paz, and Santiago Molina. We also thank Ned Cooper for his thoughtful suggestions and feedback on an earlier draft, and Leo Holton for his indispensable research assistance.

## ORCID iDs

Samuel J Bell https://orcid.org/0000-0002-9437-5449
Skyler Wang https://orcid.org/0000-0002-0639-945X

## Funding

The authors received no financial support for the research, authorship, and/or publication of this article.

## Declaration of conflicting interests

The authors declared no potential conflicts of interest with respect to the research, authorship, and/or publication of this article.

## Supplemental material

Supplemental material for this article is available online.